\documentclass[sigconf,nonacm]{acmart}
\AtBeginDocument{%
  }

\setcopyright{acmlicensed}

\usepackage{graphicx}
\usepackage{subcaption}
\usepackage[table]{xcolor}
\usepackage{enumitem}
\setlist[itemize]{leftmargin=1.2em, itemsep=2pt, topsep=2pt, parsep=0pt, partopsep=0pt}

\usepackage{algorithm2e}
\begin{document}

\title{Rethinking Point Clouds as Sequences: A Causal Next-Token Predictive Learning Framework}


\author{
Yumeng Yao\textsuperscript{1} \quad
Jingzhi Dong\textsuperscript{3} \quad
Haowen Gu\textsuperscript{1} \quad
Tao Chen\textsuperscript{1} \quad
\\
Zonghan Wu\textsuperscript{4} \quad
Xiaoshui Huang\textsuperscript{2,\ensuremath{\dagger}} \quad
Yazhou Yao\textsuperscript{1}
\\[0.5em]
\textsuperscript{1}Nanjing University of Science and Technology \quad
\textsuperscript{2}Shanghai Jiao Tong University \quad
\\
\textsuperscript{3}Hangzhou City University \quad
\textsuperscript{4}East China Normal University
\\[0.3em]
 \textsuperscript{\ensuremath{\dagger}}Corresponding author
\\[1.2em]
}




\newcommand{\algoname}{PointNTP}

\begin{abstract}
With the rapid progress of multimodal foundation models and predictive pre-training, an important open question is how to equip 3D point clouds with a pre-training paradigm that is better aligned with next-token and next-embedding learning. Existing point-cloud self-supervised methods are largely built on masked reconstruction or explicit geometric generation, and thus remain tied to input recovery rather than predictive dependency modeling.
In this paper, 
we introduce \textbf{\algoname}, which reformulates \textit{point} cloud pre-training as a \textbf{fully causal, decoder-free latent \textit{N}ext-\textit{T}oken \textit{P}rediction} problem.
Specifically, each point cloud is first partitioned into local patches and serialized into a structured 3D token sequence according to patch-center geometry. The resulting sequence is then modeled by a causal Transformer under prefix-only conditioning, and trained with a shift-based prediction objective stabilized by stop-gradient targets. This design enables the model to learn structural dependencies directly in latent space, without reconstruction decoders or explicit geometric recovery.
Extensive experiments demonstrate that the proposed \algoname~ is highly competitive across multiple downstream tasks: 
it achieves 93.8\%($\uparrow 0.5\%$), 92.6\%($\uparrow 0.3\%$), and 89.3\%($\uparrow 1.1\%$) on OBJ\_BG, OBJ\_ONLY, and PB\_T50\_RS of ScanObjectNN, respectively; obtains 85.0\%($\uparrow 0.1\%$) in Cls.mIoU on ShapeNetPart; and reaches 71.1\%  mAcc on S3DIS Area 5. Overall, decoder-free causal latent prediction provides a simple, scalable, and potentially modality-agnostic paradigm for point-cloud self-supervised learning, offering a new 3D perspective on foundation-style predictive learning for 3D data. 
\end{abstract}

\ccsdesc[500]{Computing methodologies~Computer vision}

\keywords{point cloud pre-training, self-supervised learning, causal prediction, next-token prediction, 3D representation learning, Transformer, multimedia foundation models}


\maketitle
\pagestyle{plain}

\begin{figure}[!t]
  \centering
  \includegraphics[width=0.98\columnwidth]{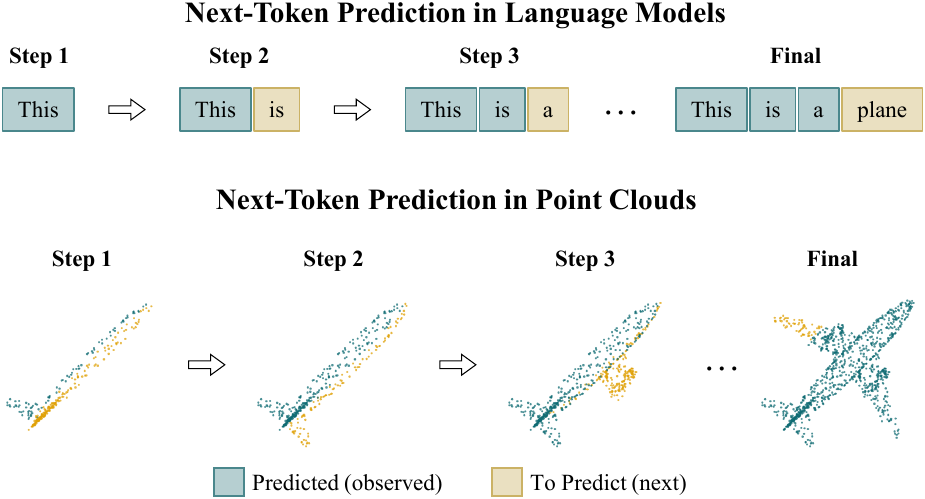}
  \caption{PointNTP: Causal Next-Token Prediction for Point Clouds. Light green indicates observed tokens, and light yellow indicates the next token to be predicted.}
  \label{fig:teaser1}
  \vspace{-1mm}
\end{figure}

\section{Introduction}

Point clouds are a fundamental 3D representation for robotics, autonomous driving, AR/VR, 3D reconstruction, digital twins, and embodied intelligence~\cite{guo2021survey}. Unlike images defined on regular grids, point clouds are inherently unordered and irregular. This has motivated dedicated architectures for direct point-set modeling, starting from PointNet~\cite{qi2017pointnet} and continuing to more expressive hierarchical and attention-based backbones~\cite{qi2017pointnetplusplus,zhao2021pointtransformer}. Since these representations underpin downstream tasks such as classification~\cite{uy2019scanobjectnn}, segmentation~\cite{armeni20163d}, and scene understanding~\cite{dai2017scannet}, learning robust, transferable, and structure-aware 3D features remains a central problem in point-cloud understanding~\cite{yu2022pointbert}.

In this context, self-supervised pre-training has become a central paradigm for point-cloud representation learning and has substantially advanced 3D classification, segmentation, and detection. Existing methods can be broadly grouped into three families. 
The first is contrastive learning, whose core idea is to enforce representation consistency across different views, augmentations, or modalities, as exemplified by PointContrast~\cite{xie2020pointcontrast}, DepthContrast~\cite{zhang2021anypointcloud}, and CrossPoint~\cite{afham2022crosspoint}. The second is reconstructive or generative learning, which learns 3D structure by recovering masked geometry, completing missing regions, or explicitly generating point clouds, as represented by OcCo~\cite{wang2021occo}, Point-BERT~\cite{yu2022pointbert}, Point-MAE~\cite{pang2022pointmae}, MaskPoint~\cite{liu2022maskpoint}, Point-M2AE~\cite{zhang2022pointm2ae}, IAE~\cite{yan2023iae}, and PointDif~\cite{zheng2024pointdif}. The third is cross-modal pre-training, which transfers complementary supervision from images, language, or rendering spaces to improve 3D representations, including ACT~\cite{dong2023act}, TAP~\cite{wang2023tap}, Ponder~\cite{huang2023ponder}, Joint-MAE~\cite{guo2023jointmae}, ULIP~\cite{gao2023ulip}, I2P-MAE~\cite{zhang2023i2pmae}, and
Point-to-Pixel Prompting (P2P)~\cite{wang2022p2p,wang2024p2p}. 
Despite their empirical success, these methods still rely, to varying degrees, on contrastive pair construction, mask tokens, reconstruction decoders, rendering branches, or additional teacher signals. In particular, many reconstruction-based pipelines depend on dedicated decoding or recovery modules, making the pre-training objective tightly coupled to input restoration rather than representation forecasting. More importantly, their supervision is typically anchored to recovering inputs or matching external targets, rather than directly modeling the conditional dependencies among 3D tokens themselves. As a result, they remain less naturally aligned with the next-token / next-embedding predictive paradigm that has emerged as a scalable and unifying principle in multimodal foundation models.

Meanwhile, foundation models in language and vision increasingly suggest that \emph{predictive learning} itself can serve as a scalable and unifying training principle. Large language models are built on next-token prediction~\cite{radford2018gpt}, while recent latent / next-embedding prediction results in vision---such as I-JEPA~\cite{assran2023ijepa} and NEPA~\cite{xu2025nepa}---
show that strong representations can be learned without heavy reconstruction heads, by directly forecasting future states instead. This naturally raises a more fundamental question: if point clouds are rethought as sequences rather than merely unordered sets, can a genuinely 3D causal next-token predictive learning framework be built? Yet this question is non-trivial. Unlike text sequences or image grids, point clouds lack a canonical token order, so ``the next token'' is not naturally defined. Recent results from PointGPT~\cite{chen2023pointgpt} and PointMamba~\cite{liang2024pointmamba} suggest that autoregressive modeling can be promising in 3D once an appropriate serialization is available; however, a point-cloud pre-training framework centered on \emph{causal next-token prediction itself}, rather than auxiliary reconstruction, auxiliary generation, or teacher distillation, remains underexplored.

Motivated by this observation, we propose \textbf{\algoname}, which reformulates point-cloud self-supervised pre-training as a fully causal, decoder-free latent next-token prediction problem. Figure~\ref{fig:teaser1} provides a conceptual overview of this idea: once a point cloud is decomposed into local patches and serialized into a structure-preserving 3D token sequence, a causal model can be trained to predict the next latent token from its prefix only. Our key idea is not to treat serialization as a peripheral engineering trick added to an existing framework, but to rethink from the objective itself whether point clouds should be modeled as sequences. Specifically, we first decompose a point cloud into local patches and construct a structured spatial serialization order based on patch centers. We then combine Center Additive PE, 3D RoPE, and a causal Transformer to model the resulting 3D token sequence under prefix-only conditioning. Finally, we adopt a shift-based latent prediction objective that predicts the next latent token solely from preceding tokens and is stabilized by a stop-gradient target.
Unlike representative methods~\cite{yu2022pointbert,pang2022pointmae,zhang2022pointm2ae,zheng2024pointdif,dong2023act},
our method neither reconstructs masked geometry nor introduces a reconstruction decoder, rendering supervision, or cross-modal teacher models; instead, it learns 3D structural dependencies directly in latent token space.

This design leads to two more fundamental innovations. First, we develop a simple latent-space next-token prediction objective that removes the need for reconstruction decoders, rendering supervision, and cross-modal teacher models, thereby offering a lighter and more unified predictive learning paradigm for point-cloud pre-training. Second, from a foundation-model perspective, our method shows that next-token predictive learning is not limited to language and vision, but can also be effectively extended to point clouds. Once a sufficiently structure-preserving 3D serialization is available, point-cloud pre-training can be naturally brought into the broader methodological framework of foundation-style predictive learning, narrowing the gap between 3D pre-training and foundation models in other modalities.

Extensive experiments validate the effectiveness of this idea. On ScanObjectNN, \algoname~ achieves 93.8, 92.6, and 89.3 on OBJ\_BG, OBJ\_ONLY, and PB\_T50\_RS, respectively. On ShapeNetPart, it obtains 85.0 / 86.2 in Cls.mIoU / Inst.mIoU. On S3DIS Area 5, it reaches 71.1 / 60.0 in mAcc / mIoU. Further ablations show that autoregressive shifting, causal masking, stop-gradient stabilization, and a causally consistent transfer interface are all critical to the final performance, while explicit center-based positional modeling, geometric coordinate transforms, and structured spatial serialization jointly support the causal predictive framework. These results suggest that the gains do not come from mechanically stacking modules, but from an internally coherent causal sequence-modeling design.

Our contributions are summarized as follows:
\begin{itemize}
    \item We propose \textbf{\algoname}, which reformulates point-cloud self-supervised pre-training as a decoder-free, fully causal latent next-token prediction problem.
    \item We transform unordered point clouds into a 3D token sequence suitable for causal modeling through structured spatial serialization, Center Additive PE, and 3D RoPE.
    \item We develop a simple latent-space next-token prediction objective that removes the need for reconstruction decoders, rendering supervision, and cross-modal teacher models, thereby establishing a lighter and more unified predictive pre-training paradigm for point clouds.
    \item From a foundation-model perspective, we show that next-token predictive learning can be effectively extended to point clouds, and verify this claim through results and systematic ablations on ScanObjectNN, ShapeNetPart, and S3DIS Area 5.
\end{itemize}

\section{Related Work}

\noindent\textbf{Self-supervised pre-training for point clouds.} Existing point-cloud pre-training methods can be broadly grouped into three families: contrastive learning~\cite{afham2022crosspoint}, reconstructive / generative learning~\cite{wang2021occo,yan2023iae}, and cross-modal or guidance-enhanced pre-training~\cite{guo2023jointmae,zhang2023i2pmae}. Contrastive methods~\cite{xie2020pointcontrast,zhang2021anypointcloud} learn invariant 3D features through cross-view or cross-modal consistency, but often depend on augmentation and pairing design~\cite{chen2023clip2scene,jing2020mvif}. Reconstructive and generative methods~\cite{pang2022pointmae,yu2022pointbert,zheng2024pointdif} learn structure through completion, masked recovery, implicit modeling, or explicit geometric generation, yet they typically rely on dedicated decoders or explicit recovery branches~\cite{liu2022maskpoint,zhang2022pointm2ae}. While effective, this line of work usually depends on mask tokens, dedicated decoders, or additional geometric recovery / generation branches, which not only increase architectural and optimization complexity, but also tie supervision closely to reconstructing missing geometry. As a result, the learned objective is often centered on input restoration fidelity rather than directly modeling the conditional dependencies needed for scalable predictive representation learning. Cross-modal and guidance-enhanced methods~\cite{dong2023act,wang2023tap} further exploit external supervision from images, language, or auxiliary objectives, but usually introduce teacher models, rendering branches, or additional alignment mechanisms~\cite{zha2025pointmode,qi2023recon}. Overall, existing point-cloud pre-training methods remain largely centered on two directions at the methodological level: decoder-heavy / explicit-recovery learning and teacher-guided transfer. In contrast, our method removes the need for reconstruction decoders, explicit geometric recovery, rendering supervision, and cross-modal teacher models, and instead performs fully causal next-token prediction directly in latent space.

\noindent\textbf{Predictive learning and the next-token / next-embedding paradigm.}
In language and vision, predictive learning has emerged as a powerful principle for representation learning. GPT~\cite{radford2018gpt} established generative pre-training through next-token prediction with a unidirectional Transformer, while Image-GPT~\cite{chen2020igpt} extended this idea to vision by autoregressively predicting discretized image tokens, showing that sequence prediction itself can learn competitive visual representations even without explicitly exploiting 2D convolutional inductive biases. Earlier contextual language models such as ELMo~\cite{peters2018deep} further suggested that predictive contextual modeling can already produce strong transferable features. More recently, predictive learning in vision has moved beyond pixel-level autoregression toward latent-state prediction: I-JEPA~\cite{assran2023ijepa}, NEPA~\cite{xu2025nepa}, and subsequent next-embedding / world-model approaches show that strong representations can be learned by predicting future latent states directly, without relying on heavy reconstruction heads. Taken together, these works suggest that predicting future tokens or embeddings can itself serve as a unified and scalable pre-training principle. Yet, compared with language and vision, this paradigm remains relatively underexplored for point clouds. Most existing point-cloud methods still formulate the task as contrastive alignment, masked recovery, or cross-modal distillation rather than as a simple and unified causal predictive objective. Our work follows this broader trajectory of predictive learning, but asks a more specific 3D question: once a sufficiently structure-preserving serialization is available for point clouds, can a fully causal, decoder-free latent next-token prediction framework be built?

\begin{figure*}[t]
  \centering
  \includegraphics[width=\textwidth]{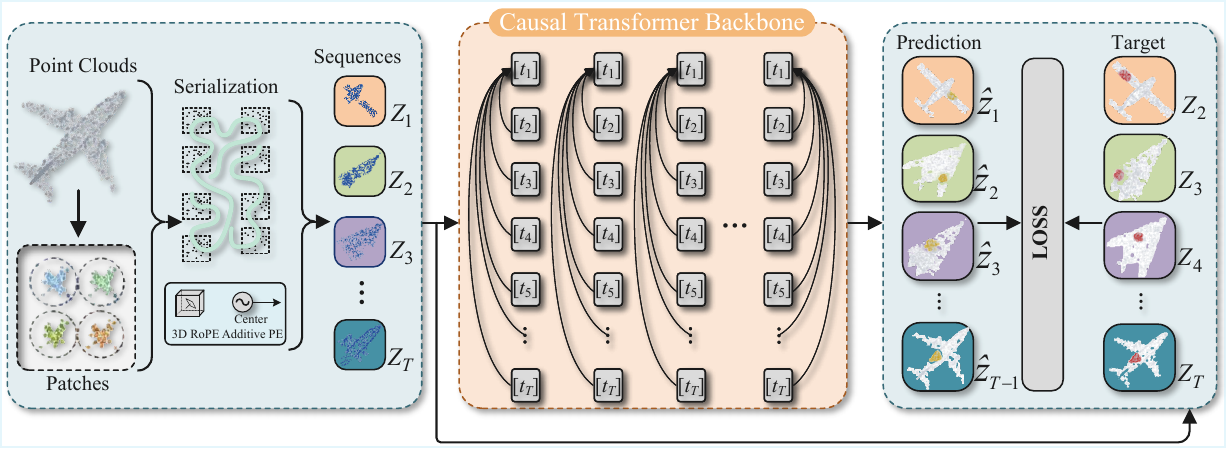}
  \vspace{-4mm}
  \caption{\textbf{Overall architecture of \algoname.} \algoname~ reformulates point-cloud self-supervised pre-training as a fully causal, decoder-free latent next-token prediction. An input point cloud is partitioned into local patches, serialized into a structured 3D token sequence based on patch-center geometry, and then processed by a causal Transformer with explicit positional modeling under strict prefix-only attention. Training uses a one-step shifted latent prediction objective with stop-gradient targets, enabling the model to learn structural dependencies directly in latent space rather than through masked reconstruction.}
  \vspace{-3mm}
  \label{fig:teaser}
\end{figure*}

\noindent\textbf{Point-cloud serialization and causal sequence modeling.}
The works most closely related to ours are those that serialize point clouds for sequential or unidirectional modeling. Since point clouds are inherently unordered sets, a prerequisite for next-token prediction or causal sequence modeling in 3D is to construct a token order that preserves local geometric continuity as much as possible. PointGPT~\cite{chen2023pointgpt} takes an important step in this direction by organizing point patches with Morton/Z-order and introducing an autoregressive generative pre-training framework for point clouds. However, PointGPT is not a minimal latent predictive learner. Its pre-training pipeline is built upon an extractor--generator Transformer decoder, dual masking, relative-direction prompts, coordinate-space patch prediction, and Chamfer-distance-based generative losses. It further includes a post-pre-training stage and continues to use generative objectives as auxiliary supervision during fine-tuning. Therefore, PointGPT is better understood as autoregressive generative pre-training for point clouds, rather than as a fully causal, decoder-free latent next-token prediction framework. From a different angle, PointMamba~\cite{liang2024pointmamba} highlights the importance of structure-preserving serialization for unidirectional modeling. It argues that Mamba-style unidirectional state-space models cannot directly operate on unordered point sets, and therefore introduces Hilbert / Trans-Hilbert scanning together with an order indicator to generate and distinguish serialized point sequences. More importantly, PointMamba explicitly formulates its pre-training as serialization-based masked modeling: during training, one serialization strategy is sampled, most tokens are masked, and the masked patches are reconstructed with a decoder-based asymmetric autoencoder. In other words, although PointMamba adopts causal scanning and sequential modeling mechanisms at the structural level, its self-supervised objective remains fundamentally MAE-like masked reconstruction rather than fully causal next-token prediction. Compared with these works, the most fundamental difference of our method does not lie in whether we ``use ordering,'' but in how the pre-training objective itself is defined. We borrow the idea of structure-preserving serialization, yet do not treat it as an isolated module plugged into a reconstruction framework. Instead, we unify structured serialization, Center Additive PE, 3D RoPE, and a causal Transformer under a fully causal, decoder-free latent next-token prediction framework. In this sense, our novelty does not come from introducing an ordering module in isolation, but from turning the modeling view that ``point clouds can be treated as sequences'' into a genuine causal predictive learning objective.

\section{Methodology}

We reformulate point-cloud self-supervised pre-training as a fully causal, decoder-free latent next-token prediction problem. The key idea is to reinterpret an unordered point cloud as a \emph{structured token sequence}, and then learn 3D representations by predicting the \emph{next latent token} from its prefix only. In contrast to reconstruction-based pre-training, our method does not recover masked coordinates, does not introduce a reconstruction decoder, and does not rely on rendering supervision or cross-modal teacher models. Instead, it directly models structural dependency in latent token space through a causal predictive objective. This formulation is conceptually aligned with next-embedding prediction in vision and world-model learning~\cite{xu2025nepa,needreamer2026}, while being instantiated here for 3D point clouds.

\subsection{Problem Setup: Rethinking Point Clouds as Sequences}

Let an input point cloud be denoted by
\begin{equation}
X=\{x_i\}_{i=1}^{N}, \qquad x_i\in\mathbb{R}^{3}.
\end{equation}
Unlike text or images, a point cloud is inherently unordered, so the notion of a ``next token'' is not naturally defined. Our starting point is therefore to convert the unordered point set into a \emph{structured 3D token sequence}.

We first partition the point cloud into local patches. Specifically, we use farthest point sampling (FPS) to select $G$ patch centers
\begin{equation}
C=\{c_g\}_{g=1}^{G}, \qquad c_g\in\mathbb{R}^{3}.
\end{equation}
Then apply KNN grouping around each center to obtain a patch set
\begin{equation}
P=\{P_g\}_{g=1}^{G}, \qquad P_g\in\mathbb{R}^{S\times 3}.
\end{equation}
Each patch is normalized by its center:
\begin{equation}
P_g \leftarrow P_g - c_g.
\end{equation}
In this way, the basic unit of pre-training is no longer an individual point, but a patch carrying local geometric semantics.

To enable causal next-token modeling, we define a serialization order $\pi$ over patch centers and reorder the patch set as
\begin{equation}
\{P_{\pi(1)}, P_{\pi(2)}, \ldots, P_{\pi(G)}\}.
\end{equation}
The sequence index here does not correspond to physical time, but rather to a \emph{pseudo-time} induced by 3D structure. Once such an order is defined, the point cloud can be viewed as a sequence, and the learning problem becomes \emph{predicting the latent token at the next serialized position using only prefix tokens}. Therefore, our ``next token'' is neither the next word nor the next pixel, but the next 3D patch embedding after structured serialization.

\subsection{Causal Next-Embedding Prediction}

Given the serialized patch set, we first encode each patch into a latent token using a shared patch encoder:
\begin{equation}
z_t = f_{\mathrm{enc}}(P_{\pi(t)}), \qquad t=1,\ldots,T,
\end{equation}
where $T=G$ is the sequence length. This yields a sequence of latent 3D tokens
\begin{equation}
\mathbf{z}_{1:T}=\{z_1,z_2,\ldots,z_T\}.
\end{equation}

We then feed the token sequence into a causal Transformer $T_{\theta}$ with a strict causal mask:
\begin{equation}
\mathbf{h}_{1:T}=T_{\theta}(\mathbf{x}_{1:T};M_{\mathrm{causal}}),
\end{equation}
where $\mathbf{x}_{1:T}$ denotes the input sequence after positional injection, and the causal mask is defined as
\begin{equation}
M_{\mathrm{causal}}(i,j)=
\begin{cases}
0, & j \le i,\\
-\infty, & j>i.
\end{cases}
\end{equation}
Thus, token $t$ attends only to its prefix, ensuring a strict prefix-conditioned prediction setup. This design aligns with recent causal next-embedding prediction in other domains~\cite{xu2025nepa,needreamer2026}.

In the current implementation, the final normalized hidden state is directly used as the prediction output:
\begin{equation}
\hat{z}_t = h_t.
\end{equation}
Training adopts a one-step shifted objective: the output at position $t$ predicts the token at position $t+1$, rather than reproducing the current token. The target branch is detached by stop-gradient~\cite{chen2021simsiam}, and the loss is defined as
\begin{equation}
\mathcal{L}_{\mathrm{next}}
=
1-\frac{1}{T-1}\sum_{t=1}^{T-1}
\left\langle
\frac{\hat{z}_t}{\|\hat{z}_t\|_2},
\frac{\mathrm{sg}(z_{t+1})}{\|z_{t+1}\|_2}
\right\rangle,
\label{eq:next_loss}
\end{equation}

where $\mathrm{sg}(\cdot)$ denotes the stop-gradient operator.

This objective consists of three key components. First, autoregressive shifting prevents the task from degenerating into trivial same-position copying. Second, causal masking enforces true prefix-only prediction. Third, stop-gradient stabilization suppresses representational collapse and makes latent prediction trainable~\cite{chen2021simsiam,xu2025nepa}. Therefore, our pre-training objective is fundamentally different from masked reconstruction: rather than recovering explicit geometry, it learns 3D structure by modeling \emph{conditional dependency between successive latent tokens}.

\begin{table*}[t]
\centering
\small
\caption{Main classification results on ScanObjectNN. We report classification accuracy (\%) on OBJ\_BG, OBJ\_ONLY, and PB\_T50\_RS, and summarize the pre-training framework and the main auxiliary module used by each method.}
\renewcommand\arraystretch{1.06}
\label{tab:scanobjectnn_main}
\resizebox{\textwidth}{!}{
\begin{tabular}{l l l l c c c}
\toprule
Methods & Reference & Pretrain Framework & Auxiliary Module & OBJ\_BG & OBJ\_ONLY & PB\_T50\_RS \\
\midrule

\multicolumn{7}{c}{\textit{Supervised Learning Only}} \\
\midrule
PointNet~\cite{qi2017pointnet} & CVPR 2017 & supervised from scratch & N/A & 73.3 & 79.2 & 68.0 \\
PointNet++~\cite{qi2017pointnetplusplus} & NeurIPS 2017 & supervised from scratch & N/A & 82.3 & 84.3 & 77.9 \\
PointCNN~\cite{li2018pointcnn} & NeurIPS 2018 & supervised from scratch & N/A & 86.1 & 85.5 & 78.5 \\
DGCNN~\cite{wang2019dgcnn} & TOG 2019 & supervised from scratch & N/A & 82.8 & 86.2 & 78.1 \\
MVTN~\cite{hamdi2021mvtn} & ICCV 2021 & multi-view supervised learning & N/A & 92.6 & 92.3 & 82.8 \\
PointMLP~\cite{ma2022pointmlp} & ICLR 2022 & supervised from scratch & N/A & -- & -- & 85.4 \\
PointNeXt~\cite{qian2022pointnext} & NeurIPS 2022 & supervised from scratch & N/A & -- & -- & 87.7 \\
P2P-RN101~\cite{wang2022p2p} & NeurIPS 2022 & supervised from scratch & N/A & -- & -- & 87.4 \\
RepSurf-U~\cite{ran2022repsurf} & CVPR 2022 & supervised from scratch & N/A & -- & -- & 84.3 \\
ADS~\cite{hong2023ads} & ICCV 2023 & supervised from scratch & N/A & -- & -- & 87.5 \\

\midrule
\multicolumn{7}{c}{\textit{Self-Supervised Pre-training with Single-Modal}} \\
\midrule
Point-BERT~\cite{yu2022pointbert} & CVPR 2022 & masked modeling & dVAE tokenizer & 87.43 & 88.12 & 83.07 \\
MaskPoint~\cite{liu2022maskpoint} & ECCV 2022 & masked modeling & transformer decoder & 89.30 & 89.70 & 84.60 \\
Point-MAE~\cite{pang2022pointmae} & ECCV 2022 & masked autoencoder & transformer decoder & 90.02 & 88.29 & 85.18 \\
Point-M2AE~\cite{zhang2022pointm2ae} & NeurIPS 2022 & masked autoencoder & transformer decoder & 91.22 & 88.81 & 86.43 \\
Point-MAE+IDPT~\cite{zha2023idpt} & ICCV 2023 & masked autoencoder  & dynamic prompt generator & 91.22 & 90.02 & 84.94 \\
Point-MAE+DAPT~\cite{zhou2024dapt} & CVPR 2024 & masked autoencoder & dynamic adapter + internal prompt & 90.88 & 90.19 & 85.08 \\
IAE (M2AE)~\cite{yan2023iae} & ICCV 2023 & implicit autoencoder & implicit decoder & 92.5 & 91.6 & 88.2 \\
PointGPT-S~\cite{chen2023pointgpt} & NeurIPS 2023 & autoregressive generation & extractor-generator transformer decoder & 91.6 & 90.0 & 86.9 \\
PointDif~\cite{zheng2024pointdif} & CVPR 2024 & diffusion-based pretraining & conditional point generator & 93.29 & 91.91 & 87.61 \\
GPM~\cite{li2024gpm} & CVPR 2024 & autoencoding + autoregressive pretraining & dVAE tokenizer & 90.2 & 90.0 & 84.8 \\
\rowcolor{gray!15}
\textbf{Ours} & -- & \textbf{fully causal autoregression} & \textbf{none} & \textbf{93.8} & \textbf{92.6} & \textbf{89.3} \\
\midrule
\multicolumn{7}{c}{\textit{Self-Supervised Pre-training with Cross-Modal}} \\
\midrule
TAP~\cite{wang2023tap} & ICCV 2023 & cross-modal generative pre-training & 2D generator & 90.36 & 89.50 & 85.67 \\
Joint-MAE~\cite{guo2023jointmae} & IJCAI 2023 & joint masked autoencoding & transformer decoder & 90.94 & 88.86 & 86.07 \\
ACT~\cite{dong2023act} & ICLR 2023 & cross-modal teacher-guided masked modeling & teacher autoencoder + transformer decoder & 93.29 & 91.91 & 88.21 \\
UniPre3D (Std. Transformer)~\cite{wang2025unipre3d} & CVPR 2025 & cross-modal Gaussian splatting pretraining & Gaussian predictor + fusion block & 92.60 & 92.08 & 87.93 \\

\bottomrule
\end{tabular}}
\vspace{-6pt}
\end{table*}

\subsection{Model Architecture}

The proposed architecture is designed to serve the above causal predictive objective as a whole, rather than as a loose combination of independent modules. It consists of four components: local patch tokenization, structured spatial serialization, explicit positional modeling, and a causal Transformer backbone.

\subsubsection{Local Patch Tokenization}

Each local patch is encoded by a PointNet-style patch encoder:
\begin{equation}
z_t=f_{\mathrm{enc}}(P_{\pi(t)})\in\mathbb{R}^{C},
\end{equation}
where $C$ is the token dimension. In our implementation, the patch encoder aggregates local geometry through pointwise convolutions and max pooling, producing a patch-level embedding with $C=384$. Since the prediction target is a patch-level latent token rather than raw coordinates, tokenization here is not merely an input preprocessing step; it defines the semantic unit on which next-token prediction is performed.

\subsubsection{Structured Spatial Serialization}

To define the next-token order, we do not use a random permutation. Instead, we serialize patches according to their centers using locality-preserving spatial scans. In the current implementation, one order

\begin{equation}
o \in \mathrm{}\big\{\texttt{hilbert}, \texttt{hilbert-trans}\}\big.
\end{equation}
is sampled during pre-training. Let $\mathcal{H}(\cdot)$ denote a 3D Hilbert encoding. The serialized order is then obtained by
\begin{equation}
\pi_o
=
\operatorname*{argsort}
\big(
\mathcal{H}_o(c_1),\ldots,\mathcal{H}_o(c_G)
\big).
\end{equation}
Here, $\mathcal{H}_{\texttt{hilbert-trans}}$ denotes a second scan derived from an axis-transformed variant of the Hilbert traversal. The use of Hilbert-style scans follows the locality-preserving serialization intuition adopted in PointMamba~\cite{liang2024pointmamba}. However, in our framework, serialization is not introduced for Mamba-style state-space modeling; instead, it is reinterpreted as an \emph{ordering prior for causal Transformer-based next-token prediction}. Without such structured serialization, a well-defined 3D next token would not exist.

\subsubsection{Order-Specific Affine Modulation}

After serialization, we apply a lightweight order-specific affine modulation to distinguish different scan orders. For each order $o$, we maintain learnable parameters $(\gamma_o,\beta_o)$ and transform the token as
\begin{equation}
\tilde{z}_t=\gamma_o\odot z_t+\beta_o.
\end{equation}
This design is closely related to the order-indicator / OrderScale mechanism used in PointMamba~\cite{liang2024pointmamba}. In our case, it does not serve state-space sequence modeling; instead, it helps the shared Transformer backbone explicitly distinguish the statistical patterns induced by different serialization orders.

\subsubsection{Explicit Positional Modeling: Center Additive PE and 3D RoPE}

Beyond the implicit prior carried by token order itself, we explicitly inject geometric positional information into the backbone.

First, we map each patch center into token space using a two-layer MLP and add it to the token:
\begin{equation}
x_t = \tilde{z}_t + \phi(c_{\pi(t)}),
\end{equation}
where $\phi:\mathbb{R}^{3}\rightarrow\mathbb{R}^{C}$ denotes Center Additive PE. This provides an absolute geometric anchor for each token.

Second, we apply 3D RoPE to the attention queries and keys. Rotary Position Embedding (RoPE) was originally introduced in RoFormer~\cite{su2024roformer} and has also been adopted in next-embedding prediction frameworks such as NEPA~\cite{xu2025nepa}. In our case, we extend it from 1D or 2D positional indexing to 3D patch-center geometry. Concretely, patch centers are first centered and radius-normalized, then cyclically assigned to $x/y/z$ frequency bands, and finally used to rotate the $Q/K$ tensors at every attention layer. As a result, Center Additive PE provides absolute geometric anchors, while 3D RoPE injects relative geometric dependency directly into attention.

\subsubsection{Causal Transformer Backbone}

The serialized and position-enhanced token sequence is processed by a pre-norm Transformer encoder with strict causal masking. In our implementation, the backbone uses 12 layers, 6 attention heads, and hidden dimension 384. During pre-training, causal attention is always preserved so that each token depends only on its prefix. As a result, the Transformer is devoted to conditional next-token modeling rather than masked reconstruction.

This point is central to our formulation. The role of the backbone is not to decode visible or masked geometry, but to model how the current 3D context predicts the next latent token. In this sense, the architecture serves the predictive objective in Eq.~(\ref{eq:next_loss}) as an integrated whole.



\section{Experiments}

This section first introduces the experimental setup and evaluation protocols. We then evaluate the transferability of \textbf{\algoname~} on real-world object classification, part segmentation, and indoor semantic segmentation. Finally, we conduct two groups of ablation studies to analyze the key causal predictive design of the method and the positional and ordering factors that support causal sequence modeling.

\subsection{Experimental Setup}

We evaluate \textbf{\algoname~} on three representative downstream tasks: real-world object classification, part segmentation, and indoor semantic segmentation. During pre-training, we use ShapeNet~\cite{chang2015shapenet} as the unlabeled source dataset. During downstream evaluation, we report results on ScanObjectNN~\cite{uy2019scanobjectnn}, ShapeNetPart~\cite{yi2016scalable}, and S3DIS~\cite{armeni20163d}. For ScanObjectNN, we report accuracy (\%) on the three standard splits, namely OBJ\_BG, OBJ\_ONLY, and PB\_T50\_RS, with PB\_T50\_RS being the most challenging one. For ShapeNetPart, we report Cls.mIoU (\%) / Inst.mIoU (\%). For S3DIS Area 5, we report mAcc (\%) / mIoU (\%).

To organize the comparisons more clearly, we group methods into three categories: Supervised Learning Only, Single-Modal Self-Supervised Pre-training, and Methods using cross-modal information or teacher models. This grouping not only positions our method against classical supervised baselines and mainstream single-modal SSL approaches, but also presents the methodological differences from reconstruction-based, teacher-guided, and cross-modal pre-training paradigms more clearly.

During pre-training, each point cloud is first decomposed into local patches and then serialized according to patch-center coordinates. The resulting token sequence is fed into a causal Transformer and optimized with a latent next-token / next-embedding prediction objective. Unless otherwise specified, all downstream fine-tuning runs follow the same protocol, so the observed performance differences mainly stem from the pre-training objective and architectural design rather than downstream optimization details.

\begin{table}[t]
\captionsetup{type=table}
\caption{Segmentation results on ShapeNetPart and S3DIS Area 5. We report Cls.mIoU (\%) / Inst.mIoU (\%) on ShapeNetPart, and mAcc  (\%) / mIoU (\%) on S3DIS Area 5.}
\label{tab:segmentation_main}
\resizebox{\columnwidth}{!}{
\begin{tabular}{lcccc}
\toprule
Methods & \multicolumn{2}{c}{Part Seg.} & \multicolumn{2}{c}{S3DIS Seg.} \\
\cmidrule(lr){2-3} \cmidrule(lr){4-5}
& Cls.mIoU & Inst.mIoU & mAcc & mIoU \\
\midrule

\multicolumn{5}{c}{\textit{Supervised Learning Only}} \\
\midrule
PointNet~\cite{qi2017pointnet} & 80.4 & 83.7 & 49.0 & 41.1 \\
PointNet++~\cite{qi2017pointnetplusplus} & 81.9 & 85.1 & 67.1 & 53.5 \\
PointCNN~\cite{li2018pointcnn} & 84.6 & 86.1 & 63.9 & 57.3 \\
DGCNN~\cite{wang2019dgcnn} & 82.3 & 85.2 & -- & -- \\
PointMLP~\cite{ma2022pointmlp} & 84.6 & 86.1 & -- & -- \\
PointNeXt-S~\cite{qian2022pointnext} & 84.4 & 86.7 & -- & -- \\

\midrule
\multicolumn{5}{c}{\textit{Self-Supervised Pre-training with Single-Modal}} \\
\midrule
Point-BERT~\cite{yu2022pointbert} & 84.1 & 85.6 & -- & -- \\
MaskPoint~\cite{liu2022maskpoint} & 84.4 & 86.0 & -- & -- \\
Point-MAE~\cite{pang2022pointmae} & 84.2 & 86.1 & 69.9 & 60.8 \\
Point-M2AE~\cite{zhang2022pointm2ae} & 84.9 & 86.5 & -- & -- \\
Point-MAE+IDPT~\cite{zha2023idpt} & 83.8 & 85.7 & -- & -- \\
Point-MAE+DAPT~\cite{zhou2024dapt} & 84.0 & 85.7 & -- & -- \\
PointGPT-S~\cite{chen2023pointgpt} & 84.1 & 86.2 & -- & -- \\
Point-FEMAE~\cite{zha2024pointfemae} & 84.9 & 86.3 & -- & -- \\
GPM~\cite{li2024gpm} & 84.2 & 85.8 & -- & -- \\
PointMamba~\cite{liang2024pointmamba} & 84.4 & 86.2 & -- & -- \\
PCP-MAE~\cite{zhang2024pcpmae} & 84.9 & 86.1 & 71.0 & 61.3 \\
Point-PQAE~\cite{zhang2025crossreconstruction} & 84.6 & 86.1 & 70.6 & \ 61.4 \\
\rowcolor{gray!15}
\textbf{Ours} & \textbf{85.0} & \textbf{86.2} & \textbf{71.1} & \textbf{60.0} \\

\midrule
\multicolumn{5}{c}{\textit{Self-Supervised Pre-training with Cross-Modal}} \\
\midrule
CrossPoint~\cite{afham2022crosspoint} & -- & 85.5 & -- & -- \\
ACT~\cite{dong2023act} & 84.7 & 86.1 & \ 71.1 & 61.2 \\
ReCon~\cite{qi2023recon} & 84.8 & \ 86.4 & -- & -- \\
\bottomrule
\end{tabular}}
\vspace{-4pt}
\end{table}

\begin{table*}[t]
\centering
\caption{Ablation study on the causal predictive design of \textbf{\algoname}. All results are reported as classification accuracy (\%) after fine-tuning on \textbf{ScanObjectNN}. Default settings are marked in gray, and failed runs are denoted as \textit{fail}.}
\label{tab:ablation_core}

\noindent\hspace*{-0.02\textwidth}%
\begin{minipage}[t]{0.60\textwidth}
\centering
\textbf{(a) Core causal predictive objective.}\par\vspace{0.5em}
\small
\setlength{\tabcolsep}{5pt}
\begin{tabular}{cccccc}
\toprule
shifting & causal masking & stop-grad & OBJ\_BG & OBJ\_ONLY & PB\_T50\_RS \\
\midrule
$\times$   & \checkmark & \checkmark & \textcolor{red}{\textit{fail}} & \textcolor{red}{\textit{fail}} & \textcolor{red}{\textit{fail}} \\
\checkmark & $\times$   & \checkmark & 91.2 & 89.2 & 86.9 \\
\checkmark & \checkmark & $\times$   & \textcolor{red}{\textit{fail}} & \textcolor{red}{\textit{fail}} & \textcolor{red}{\textit{fail}} \\
\rowcolor{gray!15}
\checkmark & \checkmark & \checkmark & \textbf{93.8} & \textbf{92.6} & \textbf{89.3} \\
\bottomrule
\end{tabular}
\end{minipage}
\hfill\hspace*{-0.1\textwidth}
\begin{minipage}[t]{0.4\textwidth}
\centering
\vspace{0.3cm}
\textbf{(b) Attention type during fine-tuning.}\par\vspace{0.5em}
\small
\setlength{\tabcolsep}{6pt}
\begin{tabular}{cccc}
\toprule
attn type & OBJ\_BG & OBJ\_ONLY & PB\_T50\_RS \\
\midrule
bidirect & 93.1 & 91.9 & 88.2 \\
\rowcolor{gray!15}
causal   & \textbf{93.8} & \textbf{92.6} & \textbf{89.3} \\
\bottomrule
\end{tabular}
\end{minipage}
\end{table*}

\subsection{Compared with Previous Results}

\noindent\textbf{Object Classification on ScanObjectNN.}
Table~\ref{tab:scanobjectnn_main} reports the main results on ScanObjectNN. Overall, \textbf{\algoname~} achieves 93.8, 92.6, and 89.3 on OBJ\_BG, OBJ\_ONLY, and PB\_T50\_RS, respectively, showing consistent advantages over the three groups of methods compared in the table, namely supervised baselines, single-modal self-supervised pre-training methods, and methods that introduce cross-modal information or teacher models. Compared with the most representative single-modal baselines, our method improves over Point-MAE~\cite{pang2022pointmae} by 3.8, 4.3, and 4.1 percentage points on OBJ\_BG, OBJ\_ONLY, and PB\_T50\_RS, respectively, and over PointGPT-S~\cite{chen2023pointgpt} by 2.2, 2.6, and 2.4 percentage points on the same three splits. These results indicate that fully causal latent next-token prediction learns strong and robust representations for real-world point-cloud classification.

More importantly, Table~\ref{tab:scanobjectnn_main} is not only a performance comparison, but also a paradigm comparison table. In addition to downstream accuracy, we summarize the Pretrain Framework and the main Auxiliary Module used during pre-training for each method. Reconstruction-based methods, Point-MAE~\cite{pang2022pointmae} and Point-M2AE~\cite{zhang2022pointm2ae}, rely on masked autoencoding with explicit reconstruction decoders. The generative method PointGPT-S~\cite{chen2023pointgpt} depends on an extractor--generator architecture. Cross-modal methods, ACT~\cite{dong2023act} and UniPre3D, further introduce teacher or image-side modules~\cite{wang2025unipre3d}. In contrast, \textbf{\algoname~} formulates point-cloud pre-training as a fully causal, decoder-free latent next-token prediction problem: the model predicts the next latent token solely from preceding tokens and learns structural dependency through a shift-based predictive loss with a stop-gradient target, without explicit geometric recovery. In this sense, the key message conveyed by Table~\ref{tab:scanobjectnn_main} lies not only in the numerical performance, but also in the distinctive combination revealed by the Pretrain Framework and Auxiliary Module columns: fully causal autoregression + none. Taken together, these results suggest that point-cloud pre-training does not have to rely on masked geometric reconstruction, diffusion-based generation, or teacher-guided cross-modal transfer; decoder-free causal latent prediction itself is already sufficient to learn strong and transferable 3D representations.

\noindent\textbf{Semantic Segmentation.}
Table~\ref{tab:segmentation_main} reports the downstream segmentation results on ShapeNetPart and S3DIS Area 5. On ShapeNetPart, \textbf{\algoname~} achieves 85.0 / 86.2 in Cls.mIoU / Inst.mIoU. On S3DIS Area 5, it reaches 71.1 / 60.0 in mAcc / mIoU. Overall, our method achieves the best Cls.mIoU in the table on ShapeNetPart and remains competitive on Inst.mIoU; On S3DIS Area 5, although it does not achieve the best mIoU, it attains competitive performance and reaches the top mAcc among the compared methods.

From a methodological perspective, these results suggest that the benefit of \textbf{\algoname~} is not restricted to global classification. Even though we abandon explicit geometric reconstruction and remove the reconstruction decoder, the learned representation still transfers effectively to dense prediction tasks that require both local detail modeling and global structural understanding. 
Compared with the representative single-modal baseline Point-MAE~\cite{pang2022pointmae}, our method improves Cls.mIoU / Inst.mIoU on ShapeNetPart by 0.8 / 0.1 percentage points. On S3DIS Area 5, it improves mAcc by 1.2 percentage points, while trailing mIoU by 0.8 percentage points. Compared with Point-PQAE~\cite{zhang2025crossreconstruction}, our method further improves Cls.mIoU / Inst.mIoU on ShapeNetPart by 0.4 / 0.1 percentage points. Compared with the cross-modal methods ACT~\cite{dong2023act} and ReCon~\cite{qi2023recon}, it remains competitive on ShapeNetPart, while achieving competitive performance against ACT on S3DIS Area 5. Overall, these results indicate that decoder-free causal latent prediction is not only effective for classification, but can also support a broader range of 3D downstream tasks.


\subsection{Ablation on the Causal Predictive Design}

To validate the key design choices of \textbf{\algoname}, we conduct ablation studies from two closely related perspectives: whether the pre-training objective must jointly include autoregressive shifting, causal masking, and stop-gradient stabilization, and whether the attention type used during downstream fine-tuning should remain consistent with the causal formulation learned during pre-training. Unless otherwise specified, all results in this section are reported after fine-tuning on ScanObjectNN. Overall, these experiments show that the effectiveness of \textbf{\algoname~} depends not only on a non-trivial causal predictive objective, but also on a causally consistent transfer interface.



\noindent\textbf{Autoregressive shifting.}
Table~\ref{tab:ablation_core}(a) shows that removing shifting leads to downstream failure. Without shifting, the objective degenerates into trivial same-position alignment, and the model no longer needs to predict the \emph{next} latent token from its prefix. As a result, the pre-training target departs from causal next-token learning, and downstream transfer can no longer converge stably.

\noindent\textbf{Causal masking.}
Removing the causal mask drops the performance to 91.2 / 89.2 / 86.9, clearly below the full model results of 93.8 / 92.6 / 89.3. This indicates that, once every token is allowed to access the full sequence bidirectionally, the model deviates from strict prefix-conditioned prediction, which weakens the consistency between causal representation learning and downstream transfer. Therefore, causal masking is not an optional training detail, but a necessary condition for the proposed pre-training objective.

\noindent\textbf{Stop-gradient.}
Following the stabilization idea of SimSiam~\cite{chen2021simsiam}, we stop gradients on the target branch to prevent collapse. Table~\ref{tab:ablation_core}(a) further shows that removing stop-gradient causes collapse during pre-training. This suggests that stabilizing the target branch is crucial for latent prediction: if the target representation and the prediction branch are allowed to co-adapt freely, the model is much more likely to degenerate into a trivial collapsed solution. Hence, stop-gradient does not merely serve as a regularizer in our framework, but acts as a key mechanism for making causal latent prediction trainable.

Taken together, Table~\ref{tab:ablation_core}(a) shows that the gain of \textbf{\algoname~} does not come from simply attaching a regression loss to a generic Transformer, but from a genuinely non-trivial causal predictive objective.

\noindent\textbf{Attention type during fine-tuning.}
Table~\ref{tab:ablation_core}(b) compares causal attention and bidirectional attention during fine-tuning. In our 3D point-cloud setting, causal attention improves performance on OBJ\_BG, OBJ\_ONLY, and PB\_T50\_RS by 0.7, 0.7, and 1.1 percentage points, respectively. This trend differs from the observation reported by NEPA~\cite{xu2025nepa}, where bidirectional fine-tuning is more favorable for 2D image classification. Our results instead suggest that, once point clouds are serialized into token sequences, preserving the causal interface during transfer helps reduce the pretrain--fine-tune mismatch. In other words, causal attention in our framework is not merely a pre-training constraint, but also a representation interface that should be retained during downstream transfer.

Overall, Table~\ref{tab:ablation_core} shows that the effectiveness of \textbf{\algoname~} arises from the synergy between a non-trivial causal predictive objective and a causally consistent transfer interface.

\subsection{Ablation on Positional and Ordering Factors}

Beyond the causal predictive objective itself, \textbf{\algoname~} also relies on a set of position- and order-related design choices, including explicit center-based positional modeling, geometric coordinate transforms, and spatial serialization strategies. To disentangle their effects, we conduct grouped ablation studies. Unless otherwise specified, all results in this section are reported as classification accuracy (\%) after fine-tuning on ScanObjectNN. Default settings are marked in gray.


\begin{table}[t]
\centering
\caption{Ablation study on positional and ordering factors in \textbf{\algoname}. All results are reported as classification accuracy (\%) after fine-tuning on \textbf{ScanObjectNN}.  Default settings are marked in gray.}
\label{tab:ablation_positional}

\begin{minipage}{\columnwidth}
\centering
\textbf{(a) Center-based positional modeling.}\par\vspace{0.4em}
\resizebox{0.95\linewidth}{!}{
\begin{tabular}{ccccc}
\toprule
Center Additive PE & 3D RoPE & OBJ\_BG & OBJ\_ONLY & PB\_T50\_RS \\
\midrule
$\times$     & $\times$     & 91.6 & 91.7 & 87.9 \\
$\times$     & $\checkmark$ & 92.4 & 91.2 & 87.9 \\
$\checkmark$ & $\times$     & 92.3 & 91.6 & 87.8 \\
\rowcolor{gray!15}
$\checkmark$ & $\checkmark$ & \textbf{93.8} & \textbf{92.6} & \textbf{89.3} \\
\bottomrule
\end{tabular}
}
\end{minipage}

\vspace{0.8em}

\begin{minipage}{\columnwidth}
\centering
\textbf{(b) Geometric transforms.}\par\vspace{0.4em}
\resizebox{0.95\linewidth}{!}{
\begin{tabular}{lccc}
\toprule
Pre-training Transform & OBJ\_BG & OBJ\_ONLY & PB\_T50\_RS \\
\midrule
None & 91.7 & 90.9 & 88.0 \\
\rowcolor{gray!15}
Rotation & \textbf{93.8} & \textbf{92.6} & \textbf{89.3} \\
Scale\&Translate & 92.4 & 91.2 & 88.0 \\
Rotation+Scale\&Translate & 92.1 & 90.9 & 88.0 \\
\bottomrule
\end{tabular}
}
\end{minipage}

\vspace{0.8em}

\begin{minipage}{\columnwidth}
\centering
\textbf{(c) Spatial serialization.}\par\vspace{0.4em}
\resizebox{0.95\linewidth}{!}{
\begin{tabular}{lccc}
\toprule
Serialization Bank & OBJ\_BG & OBJ\_ONLY & PB\_T50\_RS \\
\midrule
Random & 88.5 & 89.5 & 82.9 \\
\rowcolor{gray!15}
Hilbert+Trans-Hilbert & \textbf{93.8} & \textbf{92.6} & \textbf{89.3} \\
Z-order+Trans-Z-order & 92.9 & 91.4 & 87.5 \\
\bottomrule
\end{tabular}
}
\end{minipage}

\end{table}

\noindent\textbf{Center-based positional modeling.}
Table~\ref{tab:ablation_positional}(a) studies explicit center-based positional modeling. Using only Center Additive PE yields 92.3 / 91.6 / 87.8, whereas enabling both Center Additive PE and 3D RoPE~\cite{su2024roformer} further improves performance to 93.8 / 92.6 / 89.3, corresponding to gains of 1.5 / 1.0 / 1.5 percentage points. This indicates a clear complementarity between the two: the former provides a stable geometric anchor for each token, while the latter injects geometry-aware relative dependency directly into attention.

This ablation isolates the contribution of explicit center-based positional modeling, rather than the mere presence of positional cues. Even without the explicit patch-center channel, local relative coordinates, spatial serialization order, and causal order still provide implicit positional cues. Table~\ref{tab:ablation_positional}(a) shows that explicitly injecting patch centers still yields additional gains beyond these cues, confirming that explicit center-based positional modeling is complementary to the implicit positional priors in the framework.

\noindent\textbf{Geometric transforms.}
Table~\ref{tab:ablation_positional}(b) analyzes geometric coordinate transforms during pre-training. Rotation achieves 93.8 / 92.6 / 89.3, clearly outperforming None with 91.7 / 90.9 / 88.0, Scale\&Translate with 92.4 / 91.2 / 88.0, and Rotation + Scale \& Translate with 92.1 / 90.9 / 88.0. Relative to Scale\&Translate, Rotation improves OBJ\_BG and PB\_T50\_RS by 1.4 and 1.3 percentage points, respectively. This suggests that, for our causal sequence modeling framework, varying orientation while largely preserving global scale structure is more effective than explicitly perturbing scale and translation.

More generally, this result also suggests that the effectiveness of augmentation strategies depends on the underlying pre-training paradigm. In conventional reconstruction-based or contrastive point-cloud pre-training, scale and translation perturbations are often standard choices. In our causal next-token / next-embedding framework, however, the more beneficial strategy is to vary orientation while preserving the model's ability to capture global structure, rather than excessively perturbing geometric scale.

\noindent\textbf{Spatial serialization.}
Table~\ref{tab:ablation_positional}(c) studies spatial serialization strategies. Hilbert+Trans-Hilbert delivers the best results, namely 93.8 / 92.6 / 89.3, substantially outperforming Random with 88.5 / 89.5 / 82.9, and also exceeding Z-order+Trans-Z-order with 92.9 / 91.4 / 87.5. This shows that, in causal next-token pre-training, serialization is not merely a reordering operation; rather, it provides a key structural prior that maps 3D patch centers into a 1D autoregressive token sequence.

Importantly, we borrow the locality-preserving scanning intuition from PointMamba~\cite{liang2024pointmamba} and adopt Hilbert+Trans-Hilbert as a strong serialization prior, but our contribution does not lie in simply reusing an ordering module. Instead, we place this serialization strategy inside a fully causal latent next-token prediction framework, and jointly design it with Center Additive PE, 3D RoPE~\cite{su2024roformer}, and a causally consistent transfer interface. In this sense, the novelty of our method lies in unifying a 3D structure-preserving serialization prior with causal prediction and positional modeling, rather than mechanically stacking existing components.

Overall, Table~\ref{tab:ablation_positional} shows that the positional and ordering design of \textbf{\algoname~} is not a loose combination of independent modules, but a hierarchical system serving the same causal predictive objective: input-level geometric transforms provide coordinate variation, sequence-level structured scanning curves provide order priors, and backbone-level Center Additive PE together with 3D RoPE~\cite{su2024roformer} inject explicit positional dependency. These components jointly support the effectiveness of causal latent prediction.

\section{Conclusion}

In this paper, we presented \textbf{\algoname}, a simple and effective framework that reformulates point-cloud self-supervised pre-training as fully causal, decoder-free latent next-token prediction. By converting an unordered point cloud into a structured token sequence through local patch tokenization and spatial serialization, PointNTP learns 3D representations with explicit positional modeling and a causal Transformer, without masked reconstruction or explicit geometric recovery. Extensive experiments show strong performance across object classification, part segmentation, and indoor semantic segmentation. More importantly, our ablation results indicate that the effectiveness of PointNTP comes not from any isolated architectural trick, but from the synergy between causal predictive learning, structured serialization, and appropriate positional priors. Overall, our study suggests that properly serialized 3D point clouds can naturally benefit from predictive learning, offering a lightweight and unified path toward foundation-style pre-training for 3D representation learning.


\bibliographystyle{ACM-Reference-Format}
\bibliography{sample-base}










\end{document}